\documentclass[runningheads]{llncs}

\usepackage{eccv}

\usepackage{eccvabbrv}

\usepackage{graphicx}
\usepackage{booktabs}
\usepackage{url}            
\usepackage{booktabs}       
\usepackage{amsfonts}       
\usepackage{nicefrac}       
\usepackage{microtype}      
\usepackage{xcolor}         
\usepackage[utf8]{inputenc} 
\usepackage[T1]{fontenc}    
\usepackage{algorithm}
\usepackage{algorithmicx}
\usepackage[noend]{algpseudocode}
\algrenewcommand\alglinenumber[1]{\tiny #1:}
\usepackage{fp}
\usepackage{wrapfig,lipsum,epstopdf}
\usepackage{lipsum}
\usepackage{rotating}
\usepackage{multirow}
\usepackage{pifont}
\newcommand{\cmark}{\ding{51}}
\newcommand{\xmark}{\ding{55}}
\usepackage{array, makecell} 
\usepackage{xcolor,colortbl}
\usepackage{comment}
\usepackage{kantlipsum}
\usepackage{tabularx}
\usepackage{amsmath}
\usepackage{diagbox}
\usepackage{float}
\usepackage{caption}
\usepackage{graphicx}
\usepackage{etoolbox}
\usepackage{enumitem}
\usepackage{adjustbox}
\usepackage{arydshln} 
\usepackage{footmisc}
\setlist{topsep=0pt,noitemsep,topsep=0pt,parsep=0pt,partopsep=0pt}
\definecolor{bgreen}{rgb}{0.0, 0.5, 0.0}
\definecolor{deepskyblue}{rgb}{0.0, 0.75, 1.0}

\usepackage[accsupp]{axessibility}  

\usepackage{hyperref}

\usepackage{orcidlink}
\begin{document}
\newcommand{\papername}{Vision-Language Understanding   of Arbitrarily Long Videos}
\newcommand{\longModel}{Goldfish}
\newcommand{\shortModel}{MiniGPT4-Video}
\newcommand{\papernameAbbrev}{Goldfish:}
\newcommand{\goldfishaicon}[1]{\includegraphics[height=30pt]{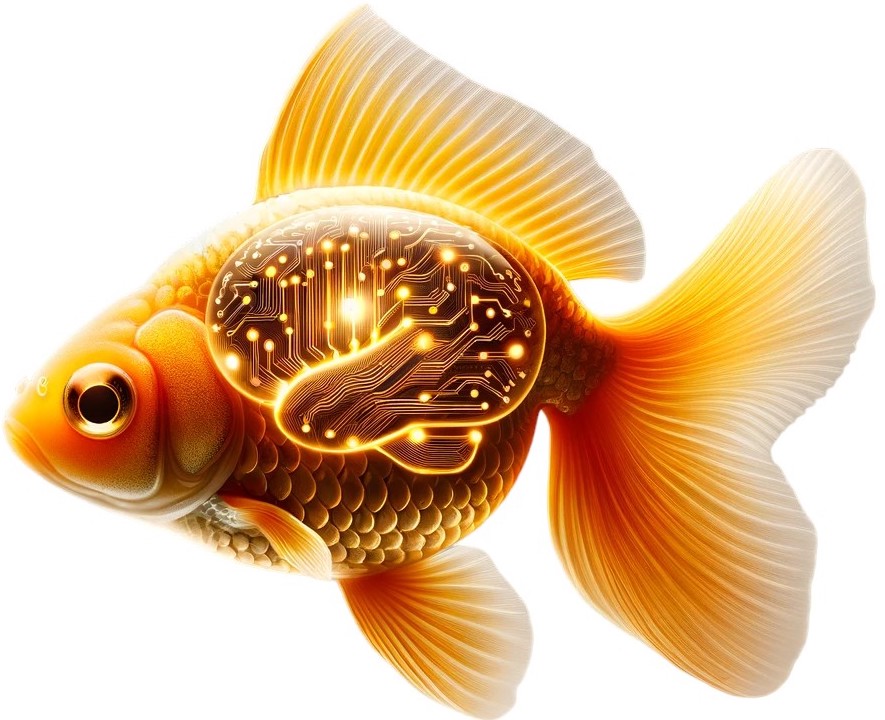}}
\newcommand{\jian}[1]{\textcolor[rgb]{0.08, 0.38, 0.74}{\textbf{Jian:} #1}}
\newcommand{\essam}[1]{\textcolor[rgb]{0.50, 0.38, 0.20}{\textbf{Essam:} #1}}
\newcommand{\mingchen}[1]{\textcolor[rgb]{0.28, 0.78, 0.74}{\textbf{Mingchen:} #1}}
\newcommand{\eslam}[1]{\textcolor{blue}{Eslam:\ #1}}
\newcommand{\jun}[1]{\textcolor[rgb]{0.58, 0.28, 0.34}{\textbf{Jun:} #1}}
\newcommand{\xiao}[1]{\textcolor{magenta}{\textbf{xiao:} #1}}
\newcommand{\kirolos}[1]{\textcolor[rgb]{0.82, 0.04, 0.9}{\textbf{kirolos:} #1}}
\newcommand{\elhoseiny}[1]{\textbf{\color{blue}[elhoseiny: #1]}}

\title{\goldfishaicon{} {\papernameAbbrev}  {\papername}}

\titlerunning{Goldfish}

\author{Kirolos Ataallah\inst{1}\orcidlink{0009-0007-0495-2171} \and
Xiaoqian Shen$^{*}$\inst{1}\orcidlink{0000-0001-6284-520X} \and
Eslam Abdelrahman\thanks{Equal contribution} \inst{1}\orcidlink{}\and
Essam Sleiman\thanks{work was done during internship in KAUST}\inst{2}\orcidlink{0000-0002-1505-6694}\and
Mingchen Zhuge\inst{1}\orcidlink{0000-0003-2561-7712}\and
Jian Ding\inst{1}\orcidlink{2222--3333-4444-5555}\and
Deyao Zhu\inst{1}\orcidlink{0000-0001-8014-7309}\and
Jürgen Schmidhuber\inst{1,3}\orcidlink{0000-0002-1468-6758}\and
Mohamed Elhoseiny\inst{1}\orcidlink{0000-0001-9659-1551}}

\authorrunning{Kirolos et al.}

\institute{King Abdullah University of Science and Technology \and Harvard University \and
The Swiss AI Lab IDSIA, USI, SUPSI }

\maketitle

\begin{abstract}
Most current LLM-based models for video understanding can process videos within minutes. However, they struggle with lengthy videos due to challenges such as ``noise and redundancy", as well as ``memory and computation" constraints. In this paper, we present \emph{\longModel}, a methodology tailored for comprehending videos of arbitrary lengths. We also introduce the TVQA-long benchmark, specifically designed to evaluate models' capabilities in understanding long videos with questions in both vision and text content.
 \emph{\longModel}~approaches these challenges with an efficient retrieval mechanism that initially gathers the top-k video clips relevant to the instruction before proceeding to provide the desired response. 
This design of the retrieval mechanism enables the \longModel~to efficiently process arbitrarily long video sequences, facilitating its application in contexts such as movies or television series. To facilitate the retrieval process, we developed \shortModel~that generates detailed descriptions for the video clips. In addressing the scarcity of benchmarks for long video evaluation, we adapted the TVQA short video benchmark for extended content analysis by aggregating questions from entire episodes, thereby shifting the evaluation from partial to full episode comprehension. We attained a 41.78\% accuracy rate on the TVQA-long benchmark, surpassing previous methods by 14.94\%. Our \shortModel~also shows exceptional performance in short video comprehension, exceeding existing state-of-the-art methods by 3.23\%, 2.03\%, 16.5\% and 23.59\% on the MSVD, MSRVTT, TGIF,and TVQA short video benchmarks, respectively.  These results indicate that our models have significant improvements in both long and short-video understanding.Our models and code have been
made publicly available \href{https://vision-cair.github.io/Goldfish_website/}{Goldfish}.
\end{abstract}

\section{Introduction}
\label{sec:intro}
\begin{figure}
    \captionsetup{font=scriptsize}
    \centering
    \scalebox{0.70}{
    \includegraphics[width=\linewidth]{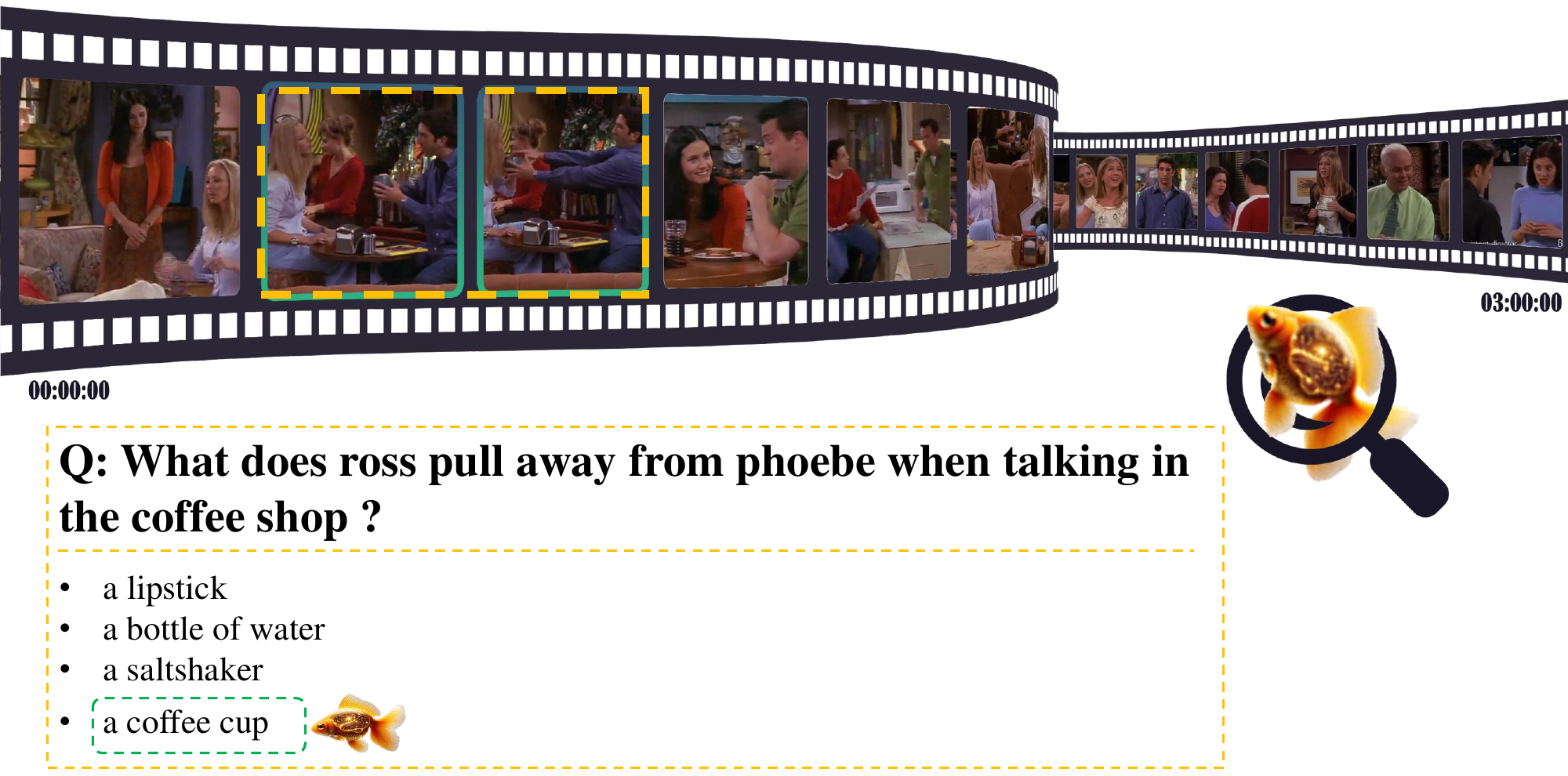}
    }
    \caption{GoldFish Model: A long-video model capable of handling lengthy videos by filtering out noisy information and focusing on the most relevant content to accurately answer questions.}
    \label{fig:teaser}
\end{figure}

The complex and detailed nature of videos provides deep insight, making them crucial for understanding and interacting with the visual world. Recent advances in large vision language models (VLMs) have progressed from image to video-centric multimodal dialogue system~\cite{video-llava, moviechat, llamavid, video-llama, li2024videochat, videochatgpt}, enabling these models to process and respond to inputs comprising a video, a user query and, optionally, video subtitles. Despite the progress in adapting VLMs for video, most of the previous works~\cite{video-llava,videochatgpt,video-llama,li2024videochat} focus on understanding short videos (in minutes) and struggle to deal with long videos. Recent approaches attempted to address this limitation. For example, MovieChat~\cite{moviechat}use memory consolidation module and LLaMa-Vid~\cite{llamavid} compress image representations into fewer tokens. These strategies improve the capacity to handle larger context windows, enabling these models to process significantly longer videos. However, this compression results in the loss of spatial and temporal visual details and leads to unsatisfactory performance in the understanding of long videos (see Tab.~\ref{tab:benchmark_longvideo}). We question: \textit{what factors contribute to the increased difficulty in understanding long videos compared to short videos?} We approach this question by identifying several challenges: 
\begin{itemize}
    \item \textbf{Noise and Redundancy}: As demonstrated in the ``needle in a haystack'' test~\cite{Doe2021Needle} in the NLP domain, LLMs tend to overlook valuable information within overly extensive contexts. Similarly, long videos often contain irrelevant or redundant information, Making it challenging for the current video-centric LLM to extract meaningful content, especially with a collapsed spatial as well as temporal resolution.
    \item \textbf{Computational and Memory Complexity}: The longer the video, the greater the computational and memory costs required for processing. Current video-centric Large Language Models (LLMs)~\cite{videochatgpt, video-llava, video-llama} inherently always have a limitation on the maximum length of videos that they are capable of processing.
    \item \textbf{Lacking Effective Benchmarks for long video understanding}: Existing benchmarks for long videos, such as
    LLama-Vid~\cite{llamavid}, primarily generate questions by feeding movie summaries and scripts into a language model, omitting visual data. This approach leads to questions that are text-centric and may be answerable without needing access to the visual content.
\end{itemize}

To address the challenges of \textit{noise and redudancy} and \textit{computational and memory costs}, we argue that \textit{ accurate identification of video clips relevant to queries is a crucial aspect in understanding long videos}. 
We propose \longModel, a framework for understanding videos of arbitrary lengths; see Fig.~\ref{fig:teaser}. \longModel~addresses these issues by incorporating a retrieval mechanism that selects the top-k relevant video clips before responding to queries. Specifically, \longModel~ segments long videos into shorter clips, applies a \textit{Video Descriptor} module to each clip to generate a detailed description of each video clip, and then executes \textit{retrieval module} by comparing the similarities in the text domain between the query text embeddings and the detailed description text embeddings. Following this, the query and corresponding summaries are forwarded to an \textit{answer module} to formulate responses. 
The Video Descriptor module is actually a short video model (\shortModel), which extends MiniGPT-v2~\cite{minigptv2}'s architecture to encode not just a single image, but multiple frames with their aligned subtitles. We map the frame tokens through a linear layer to language tokens. Following this, we tokenize the user query and the video subtitles, then introduce both lists of tokens to the LLM, this model not used by zero shot image level but trained on three stages by using video data to enhancing our model's ability to interpret and respond to video content, and this is one of our contribution as we achieved SOTA results for short video benchmarks.
In addressing the challenge of \textit{Lacking Effective Benchmarks for long video understanding}, we adapted the TVQA short video benchmark for extended content analysis by aggregating questions from entire episodes, thereby shifting the evaluation from partial to full episode comprehension. We extensively evaluated the proposed \longModel~on previous video benchmarks and our proposed long video benchmark and demonstrated superiority for long video understanding. For example, \longModel~surpasses the competitive concurrent work LLaMA-VID model~\cite{llamavid} by about 15\% in accuracy. The proposed \shortModel~also outperforms existing state-of-the-art methods by 3.23\%, 2.03\%, 16.5\% and 6.43\% on the MSVD, MSRVTT, TGIF,and TVQA short video benchmarks.
Our contributions can be summarized as follows:
\begin{itemize}
    \item We developed the \longModel~framework for long video understanding, which eased the challenges of long video understanding by introducing a retrieval design. Only top-k relevant video clips are used to answer the questions. \textit{While most previous works can only perform couple of minutes videos, \longModel~can efficiently process arbitrarily long videos.}
    \item We proposed a new TVQA-long benchmark for long video understanding. Compared to the previous long video benchmarks, TVQA-ong benchmark requires the model to understand both the visual and text content.
    \item We developed \shortModel, which extends VLM to process from single image to multiple frames. By converting frame tokens to language tokens and incorporating the user's query, we improved the model content understanding by training it for 3 stages by using video data. \shortModel~can function both as a component for detailed video descriptor within \longModel~and as an independent model for short video tasks.
    \item Our proposed \longModel~is adept at processing long video understanding, which is verified by achieving SoTA experimental results on 4 long video benchmarks, including LLama-Vid, MovieChat, Movie QA and TVQA with only the vision content and achieved SOTA results with vision and subtitles with zeroshot evaluation on TVQA as TVQA is the only benchmark can be used for zeroshot evaluation because the other models trained on the movies datasets. Apart from the long video understanding, our \shortModel~also outperformed other methods on 5 short video benchmarks, including Video ChatGPT benchmark, MSVD, MSRVTT, TGIF, and TVQA.
    
\end{itemize}

\section{Related Work}
\subsection{LLM-Based Short Video Understanding}
Recently, vision-language models such as 
Video-LLaMA~\cite{video-llama} and VideoChat~\cite{li2024videochat} extend the BLIP-2~\cite{blip2} architecture for video embedding extraction and both employ two streams for audio and visual signals. Video-LLaMA employs a Video Q-Former and an Audio Q-Former for the two streams, while VideoChat has a video embedder and a perception toolkit for captioning, tags, etc.
On the other hand, Video-ChatGPT~\cite{videochatgpt} leverages a single stream where the architecture first encodes each frame and then has a spatial and temporal pooling process that is finally mapped to an LLM with a linear layer. Video LLaVA~\cite{video-llava} takes advantage of the LanguageBind module to map both image and video inputs to the same embedding space. 
\subsection{LLM-Based Long Video Understanding}
Understanding long videos, such as movies or TV series that exceed two hours in duration, poses significant challenges (as we discussed in Sec.~\ref{sec:intro}) for current video-centric multimodal dialogue systems~\cite{video-llava,videochatgpt,video-llama}.
 Recent MovieChat~\cite{moviechat} attempts to address this problem with a memory module containing both \textit{long-term} and \textit{short-term} memory. Short-term memory consists of dense frame-wise encodings that are managed in a FIFO (First In, First Out) queue. When short-term memory is full, the contents are sent to a memory consolidation module, which combines adjacent embeddings by merging similar ones, and then stores them in long-term memory. However, the memory mechanism of this work struggles to capture meaningful information relevant to specific tasks.
A concurrent work LLaMA-VID~\cite{llamavid} builds a more efficient method by representing each frame with only two tokens, namely context token and content token. These two methods compress the input frame embeddings, increasing the number of frames fitting into the model context window.\\

Both MovieChat~\cite{moviechat} and LLaMA-VID~\cite{llamavid} have addressed the \textit{computation and memory} challenge to an extent by compressing visual features. However, their approach of using features from the entire video to predict answers has led them to face issues with \textit{noise and redundancy challenge}. In \longModel, we introduce a retrieval-based framework that utilizes only the top-k relevant video clips for question answering. This retrieval approach mitigates both challenges and enables efficient processing of long videos.

\subsection{Retrieval Systems}
LLMs have recently shown promising capabilities in a wide range of different tasks, however, face challenges such as hallucination, when a model outputs a nonsensical or incorrect output typically on queries that extend outside of its training data. Retrieval-Augmented Generation (RAG) is a technique where an LLM leverages an external knowledge base through a retrieval mechanism, mitigating hallucinations while storing long context. There are multiple RAG variations introduced for language retrieval~\cite{lewis2021retrievalaugmented, khandelwal2020generalization, ram2023incontext, li2024chainofknowledge, li2024chainofknowledge, wang2023selfknowledge, khattab2023demonstratesearchpredict, liang2023intergen, gu2018search, zhang2018guiding, peng2019text, weston2018retrieve, wu2018response,zhuge2021kaleido} and recently have been translated for image retrieval as well ~\cite{lin2022retrieval, karpukhin2020dense, chen2022murag}. Most recently there has also been some work in video retrieval ~\cite{le2022vgnmn, whitehead2018incorporating}, however, none of these methods can do robust, long-video retrieval. We draw inspiration from these works and develop a retrieval system in the domain of video-centric LLM for long-video retrieval. 

\section{{\longModel}}
\subsection{Retrieval-based Long Video Understanding}
\begin{figure}[t!]
    \captionsetup{font=scriptsize}
    \centering
    \includegraphics[width=\linewidth]{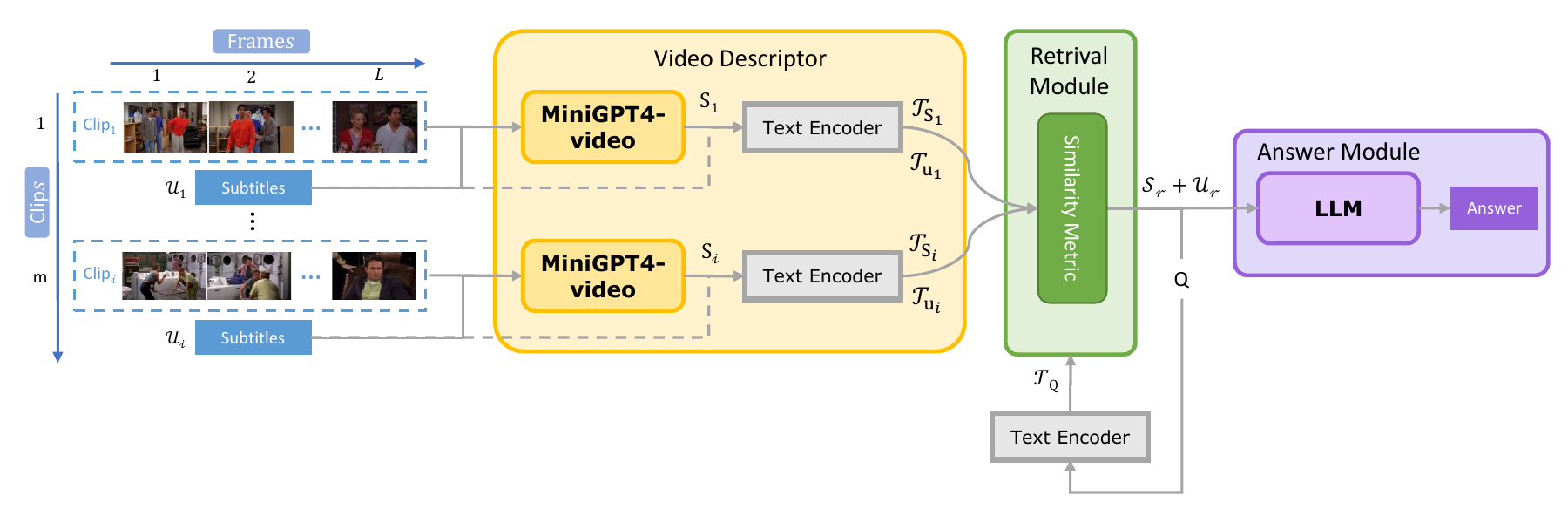}
    \caption{\textbf{Goldfish framework},First break down the long video into clips, then encode them in Video Descriptor according to their timing and corresponding subtitles, then encode the use query and retrieve the most related clips in the retrieval module, and finally send the top-K clips information to the answer module to get the final answer.}
    \label{fig:long}
    \vspace{-2em}
\end{figure}

To understand long videos that exceed the context of a normal video large language model, we introduce a three-part system: (1) \textit{Video Descriptor} empowered by a \shortModel~model and a text encoder, (2) similarity-based \textit{Retrieval Module}, and (3) \textit{Answer Module}. An overview of our system is demonstrated in Fig.~\ref{fig:long}.
Our system works as follows.
Firstly, in our Video Descriptor,we break the long video down into smaller clips, with each clip limited by a maximum number of frames that can be supported by our \shortModel~context length (4K). Then, \shortModel~provides a concise detailed summary for each clip, which is further encoded to an embedding by a text encoder.
Given a user query encoded to an embedding by the same text encoder, our \textit{Retrieval Module} retrieves the most related $k$ clips from the long video and sends them to the \textit{Answer Module} to formulate an answer to the query.

\noindent\textbf{Video Descriptor.} 
Our Video Descriptor breaks down lengthy videos into multiple non-overlapped short clips, each accompanied by textual descriptions and corresponding embeddings for the Retrieval Module. The input for the Video Descriptor is a sequence of frames, denoted as $V = \{v_1, v_2, \ldots, v_T\}$, where $v_i\in \mathbb{R}^{3\times H\times W}$ represents the $i$-th frame, and $T$ is the sequence's length. These frames are then grouped into $m$ chunks, with each chunk represented as $C_k, k\in [1,m]$, comprising at most $L$ consecutive frames $v_{k,j}$ from the video $V$, where $v_{k,j}$ signifies the $j$-th frame within the $k$-th chunk. Here, $L$ is determined by the maximum number of frames that can be accommodated within the context window of our \shortModel~introduced later. Consequently, the video can be represented as a sequence of clips: $V = \{C_1, C_2, \ldots, C_m\}=\{(v_{1,1},...,v_{1,L}),(v_{2,1},...,v_{2,L}),...,(v_{m,1},...,v_{m,L})\}$.

\begin{figure}[t!]
    \captionsetup{font=scriptsize}
    \centering
    \scalebox{0.80}{
    \includegraphics[width=1\linewidth]{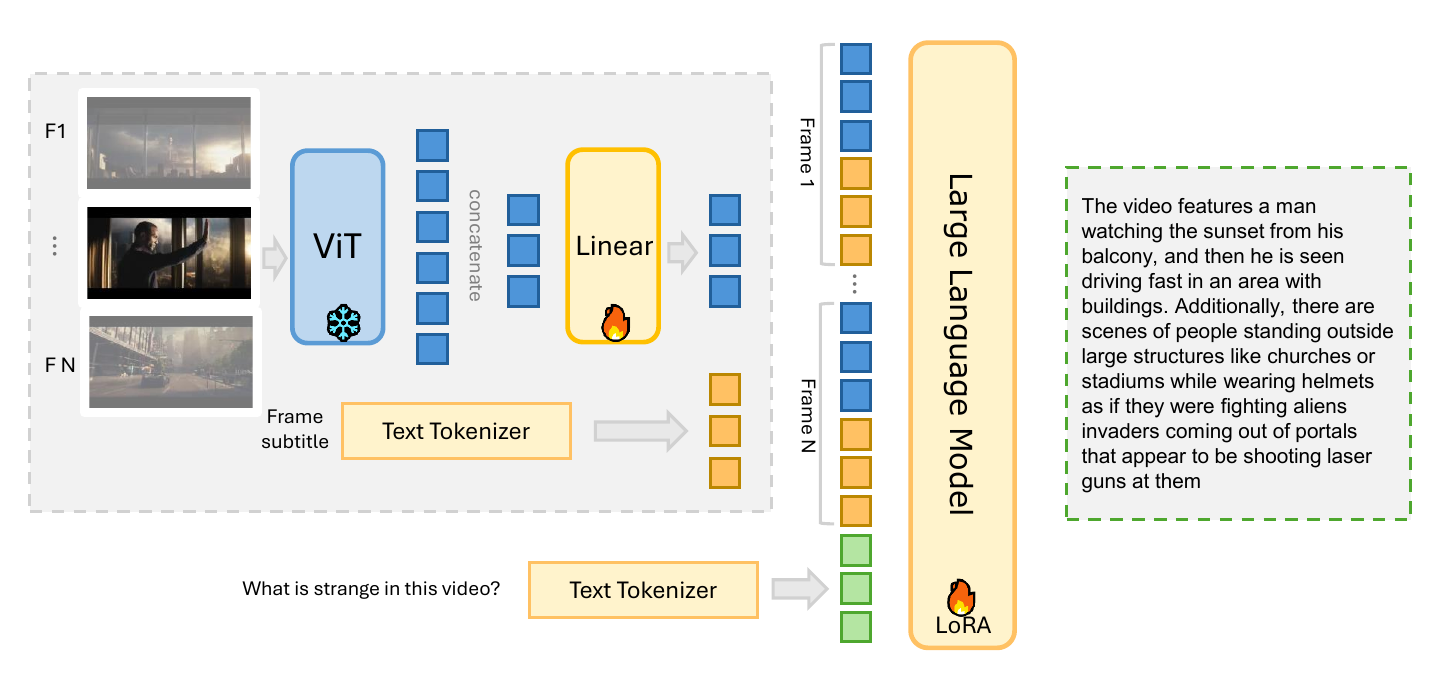}
    }
    \caption{MiniGPT4-video architecture: For each frame, we use EVA-CLIP to get the visual tokens and concatenate each adjacent visual token into a singular token then convert these tokens to the language model space using a linear layer and get the language token from LLM tokenizer.
    Concatenate both the visual and subtitle text tokens together and do this for all the sampled frames and appending the instruction tokens at the end of the input sequence.
    }
    \label{fig:minigpt4_video}
\vspace{-2em}
\end{figure}

We employ our short video model (\shortModel) to handle the processing and generation of descriptions for each video clip. Drawing from existing LLM-based vision-language models~\cite{minigptv2, llava, blip2}, we adapt this framework to the video domain, resulting in our \shortModel~model. The architecture of this model is illustrated in Fig.~\ref{fig:minigpt4_video}. For the video encoding stage, we utilize EVA-CLIP\cite{sun2023eva}, integrating a projection layer to map visual features from the vision space to the text space of the LLM. To optimize the contextual capabilities of the LLM, we condense every four adjacent visual tokens into a single token, effectively reducing the token count per image by 75\%, from 256 to 64 similar as \cite{minigptv2}. Through training, the LLM learns to process these video frame features, generating comprehensive clip descriptions $S_1, S_2, S_m$ for each clip essential for conducting visual question-answering tasks in the vision-language domain.

After generating descriptions for the video clips, we proceed to encode them along with their respective subtitles using a text encoder. The set of encoded descriptions is defined as: $\{T_{s_1}, T_{s_2}, ..., T_{s_m}\}$, and the encoded corresponding subtitles $\{u_1,u_2,...u_m\}$ are defined as $\{T_{u_1}, T_{u_2}, ..., T_{u_m}\}$, where $T_{u_i}, T_{s_i} \in \mathbb{R}^{d}, i\in [1,m]$, and $d$ is the dimensionality of text encoder space. Specifically, we employ OpenAI's \texttt{text-embedding-3-small}\cite{openaiembedding} model as our chosen text encoder based on table \ref{tab:text_encoder_ablation} in section 4.4 .

\noindent\textbf{Retrieval Module.} The Retrieval Module plays a crucial role in identifying video clips most pertinent to a user query, leveraging the pre-processed clip embeddings from the Video Descriptor. Upon receiving a user query $Q$, we initially encode it using the text encoder, resulting in the embedding $T_Q \in \mathbb{R}^{d}$. Subsequently, we compute its cosine similarities with each candidate key $K_i$ from the embeddings set of the clip descriptions and subtitles with $K_i \in \{T_{u_1}, T_{u_2}, ..., T_{u_m}, T_{s_1}, T_{s_2}, ..., T_{s_m}\}$ via $\frac{\bf K_i\cdot \bf T_Q}{|\bf K_i| |\bf T_Q|}$. Next, we select the Top-$k$ similarity scores and retrieve the corresponding descriptions or subtitle indexes, effectively eliminating irrelevant clips from the long video.

\noindent\textbf{Answer module.} 
In the final stage, we provide the original user query along with our retrieved clip descriptions (and subtitles, if available) as a context to our answer module, which generates the ultimate query response. For this purpose, we utilize Llama2-chat~\cite{llama2} as our chosen Answer module instead of \shortModel~in the text tasks. see the supplementary for more details and ablations.

\subsection{Training Pipeline}
\label{sec:training_pipeline}
\noindent\textbf{Large-scale image-text pair pretraining.} In the first stage, we train a linear layer, similar as~\cite{zhu2023minigpt}, which projects the visual feature encoded by the vision encoder (EVA-CLIP~\cite{sun2023eva}) to the LLM's text space with captioning loss. We leverage a combined image captioning dataset that includes images from LAION~\cite{schuhmann2021laion400m}, Conceptual Captions~\cite{ sharma2018conceptual}, and SBU~\cite{ordonez2011im2text} to align the visual feature with LLM's input space.
To efficiently utilize the context length of LLM for video, we concatenate every four neighboring visual tokens into a single token, reducing the number of tokens per image by 75\% from 256 to 64 same as in \cite{minigptv2}.

\noindent \textbf{Large-scale video-text pair pretraining.} In the second stage, we enable the model to understand short videos by taking multiple frames as input. Specifically, we sample a maximum of 45 frames from each short video. During this stage, we use the predefined prompts in the following template:\\\\
\textit{<s>[INST]<Img><FrameFeature\_1><Sub><Subtitle text\_1>... <Img> <FrameFeature\_N><Sub><Subtitle text\_N><Instruction><\//INST>}\\\\
where $N\leq 45$. In this prompt, each \textit{<FrameFeature>} is replaced by the sampled video frame encoded by the vision backbone. The \textit{<Subtitle text>} represents the subtitle for the corresponding frame if applicable, and \textit{<Instruction>} represents a randomly sampled instruction from our predefined instruction set containing variant forms of instruction, such as \textit{``Briefly describe this video''}. We use combined video captioning data incorporating CMD~\cite{bain2020condensed} and WebVid~\cite{webvid} for large-scale video captioning training.

\noindent \textbf{Video question answering instruction finetuning.} In this phase, we adopt the same training strategy implemented in the second stage but focus on leveraging high-quality video-question-answering datasets for instruction fine-tuning. This fine-tuning stage helps to enhance the model's ability to interpret the input video and generate precise responses to the corresponding questions.
The template is the same as the second stage with \textit{<Instruction>} replaced by general questions as mentioned in the Video-ChatGPT~\cite{videochatgpt} dataset.
\section{Experiments}

\subsection{Datasets}
\subsubsection{Training Datasets}

The Condensed Movies Video Captions dataset (CMD)~\cite{bain2020condensed} includes around 15,938 videos, with lengths between one to two minutes. However, CMD's captions are of limited quality, featuring an average sentence length of 14 words so we used it in the pre-training stage.

The Webvid dataset~\cite{webvid} contains two million videos. For our purposes, we've filtered 42K from this dataset to match CMD's video duration range, focusing on videos lasting one to two minutes and also used this dataset in the pre-training dataset.

The Video Instruction Dataset~\cite{maaz2023video} offers 100K question-answer pairs across 13,224 videos, distinguished by its high-quality annotations. Questions come with detailed answers, averaging 57 words per sentence. This data set spans various types of questions, including video summarization-based and description-based QAs that delve into spatial, temporal, relationships, and reasoning aspects, as well as creative or generative QAs.

\subsubsection{Short Benchmarks}
Our \shortModel~is tested with Video ChatGPT benchmark five skills and with open-ended and MCQ video-question answering benchmarks. 
The Video ChatGPT benchmark~\cite{maaz2023video}, utilizing the ActivityNet-200 dataset~\cite{caba2015activitynet}, is designed to test video-based conversation models on text generation, focusing on five critical dimensions:
1) Correctness of Information: Verifies the generated text's accuracy with video content to avoid errors or misinformation.
2) Detail Orientation: Assesses the responses for thoroughness and detail, ensuring coverage of essential video elements and inclusion of specific, rather than broad, information.
3) Contextual Understanding: Gauges the model's grasp of video context, ensuring responses are contextually appropriate.
4) Temporal Understanding: Checks the model's perception of event sequences within the video.
5) Consistency: Tests output reliability through similar question comparisons.
For open-ended questions, model performance is measured using established datasets like MSRVTT-QA~\cite{xu2017video}, MSVDQA~\cite{xu2017video}, TGIF-QA FrameQA~\cite{jang2017tgifqa}, and ActivityNet-QA~\cite{yu2019activitynetqa}. \\
For multi-choice question assessments utilize the TVQA dataset~\cite{lei2019tvqa}, based on popular six TV shows, with a validation set of 15,253 QA pairs for evaluation.

\subsubsection{Long Benchmarks}
We have conducted comprehensive evaluations on three extensive and demanding long video benchmarks: Movie-QA~\cite{movieqa}, LLama-vid~\cite{llamavid}, and Movie Chat~\cite{moviechat}. Additionally, we adapted the short video benchmark TVQA for long video analysis.

For Movie-QA~\cite{movieqa}, we assessed the overlapped movies between Movie-QA and MovieNet \cite{huang2020movienet} because Movie-QA videos is not avaialble and it is only short clips,we ended up with 30 overlapped movies from the validation set, each lasting between 1 and 2 hours. The new validation subset encompasses 1,081 questions, primarily based on movie plot.

The LLama-vid~\cite{llamavid} dataset features QA pairs focusing on three domains: video summary (1k), movie plot (4k), and detailed reasoning (4k). The absence of category labels prompted us to employ GPT-4 for the classification of the questions, dividing them into two types : general questions (covering plot and reasoning) and summary questions. Due to the original dataset's training-only designation and lack of a validation set, we created a balanced validation set of 10 \% of the full data comprising 800 general questions and 100 summary questions, focused solely on textual content.

Movie Chat~\cite{moviechat} includes 1,000 meticulously selected video clips from a variety of movies and TV shows, accompanied by 14,000 manual annotations. These videos span 15 major genres and feature a comprehensive dense caption, three global mode QA pairs, and ten breakpoint-mode QA pairs with precise timestamps. The collection predominantly consists of videos lasting between 10K to 12K frames, with 14.6\% exceeding this range and only 8.6\% falling short, categorized exclusively as visual content.\\ we evaluated on only the available released training data which is about 10 \% of the data because the test data not released while implementing this project.

Furthermore, we introduce an enhanced benchmark based on TVQA, comprising a validation set with 15,253 QA pairs derived from 842 episodes, addressing both textual and visual queries. Originally focused on short videos with 1-minute clips, we have expanded the scope to incorporate entire episodes into the assessment, regardless of the specific video segment to which the question pertains. This adjustment, termed TVQA-Long, significantly increases the difficulty by requiring the analysis of the complete video content to locate the answers. This adjustment facilitates the measurement of retrieval accuracy, as the ground truth clip for each question is known.

\subsubsection{Evaluation Metrics}.
For open ended questions it is hard to evaluate the output of the llm with the ground truth, so following videochatGPT evaluation~\cite{maaz2023video} we employed GPT-3.5 turbo to compare between the generated results and the ground truth. We used the same prompt as videochatgpt~\cite{maaz2023video} to have a fair comparison with their results.

\subsection{Ablation Studies}

\subsubsection{Retrieval Importance.}
The retrieval system is one of our core contributions, thus, before ablating its inner design we conduct a simple experiment to demonstrate its importance.
To this end, given a long video as an input, we directly fed a sampled version of it.
More specifically, we downsample the input video by sampling 45 frames to fit the context length of our \shortModel~model, then fed it directly to our architecture as one clip.
This could be seen as a vanilla approach to process a long video with our \shortModel.
To avoid the huge information loss that will be caused by this vanilla approach we propose our retrieval module, that given N clips it will automatically retrieve the Top-K clips that are related to the fed question $Q$.
The performance of our model without the retrieval module is close to random, with an accuracy of approximately 25.07\%. However, when the retrieval module is incorporated, the accuracy significantly improves, rising to 41.78\% on the TVQA-Long benchmark. Notably, the TVQA-Long benchmark consists of 5 options per question, resulting in a random accuracy baseline of 20\%.

\begin{figure}[t!]
    \captionsetup{font=scriptsize}
    \centering
        \begin{minipage}{.45\linewidth}
            \centering
            \captionof{table}{Ablation study of the retrieval inputs. The reported numbers are the retrieval accuracy on the TVQA-Long, TVR-Text, and TVR-Vision. \& means ``and'' while | indicates ``or''.}
            \label{tab:Retrieval_input_ablation}
            \scalebox{0.6}{
            \begin{tabular}{cccc}
                \toprule
                \textbf{Retrieval I/P} & \textbf{TVQA}& \multirow{2}{*}{\makecell{\textbf{TVR}\\\textbf{Text}}} & \multirow{2}{*}{\makecell{\textbf{TVR}\\\textbf{Vision}}}\\
                \- & \- & \- & \- \\
                \hline
                Subtitles & 39.7 & 66.4 & 48.4\\
                Summary & 12.1 & 41.2& 51.2\\
                Subtitles \& Summary & 36.9 & 64.3& 46.7\\
                Subtitles | Summary & 39.5 & 67.2& 50.8\\
              \bottomrule
            \end{tabular}
            }
        \end{minipage}
        \begin{minipage}{.45\linewidth}
            \centering
            \captionof{table}{Ablation study on the text encoder models}
            \label{tab:text_encoder_ablation}
            \scalebox{0.6}{
            \begin{tabular}{ccc}
                \toprule
                \textbf{Text Encoder} & \multirow{2}{*}{\makecell{\textbf{Retrieval}\\\textbf{Acc.}}} & \multirow{2}{*}{\makecell{\textbf{Overall}\\\textbf{Acc.}}} \\
                \- & \- & \- \\
                \hline
                bert-base-nli-mean-tokens ~\cite{sbert} & 19.0 & 28.4\\
                paraphrase-MiniLM-L6-v2 ~\cite{sbert}& 31.9 & 38.03\\
                all-mpnet-base-v2~\cite{sbert} & 32.5  & 38.33\\
                OpenAI-text-embedding-3-small ~\cite{openaiembedding}& 46.6 & 41.78\\
              \bottomrule
            \end{tabular}
            }
        \end{minipage}
\end{figure}

\subsubsection{Retrieval Inner Design.}
After demonstrating the importance of retrieval design, we
ablate each design choice to implement an efficient retrieval system.
For each clip $i$, given a question $Q$, the subtitles and the summaries embedding, termed $E^{i}_{sub}$ and $E^{i}_{sum}$, respectively, we need to determine what is the best way to retrieve the corresponding clip to the input question.
To this end, we explored four possible approaches:
1) Using only $E^{i}_{sub}$.
2) Using only $E^{i}_{sum}$.
3) Concatenate both embedding $E^{i}_{sub}$ and $E^{i}_{sum}$, namely ``and'' approach.
4) Treat each type separately, namely ``or'' approach. For instance, if we have 20 clips, then we will feed 40 embeddings, each 20 representing the $E^{i}_{sub}$ and $E^{i}_{sum}$ separately.
As shown in Table \ref{tab:Retrieval_input_ablation}, on the TVQA dataset the summary do not add any value, which could be interpreted as our generated summary is unrepresentative.
However, other interpretation is that, the questions provided in the TVQA dataset mainly rely on the text clues not the vision ones.
To support this claim and to truly assess our generated summaries, we exploit the TVR dataset \cite{lei2020tvrlargescaledatasetvideosubtitle} which is another dataset of the same videos but different annotations that used in moment retrieval tasks and this dataset has a good prosperity that the descriptions in it is labeled as text descriptions, vision descriptions and text plus vision descriptions, so based on the description type, whether it is based on the visual clues or text.
As shown in Table \ref{tab:Retrieval_input_ablation}, on the TVR-Vision, the summary achieves the best performance, which show the high quality of our generated summaries via our short video model (\shortModel)

\subsubsection{Text Encoder.}
As shown in Figure \ref{fig:long}, the input subtitles and the generated summaries are encoded using a text-encoder to generate $E^{i}_{sub}$ and $E^{i}_{sum}$, respectively.
Table \ref{tab:text_encoder_ablation} shows the impact of the text encoder on the retrieval accuracy and the overall accuracy, where the better retrieval is linearly correlated with the overall accuracy of the long-video model.

\subsubsection{Answer Module.}
\begin{figure}[t!]
\captionsetup{font=scriptsize}
\begin{minipage}[c]{.7\linewidth}
    \vspace{0pt}
    \begin{center}
        \includegraphics[width=0.8\linewidth]{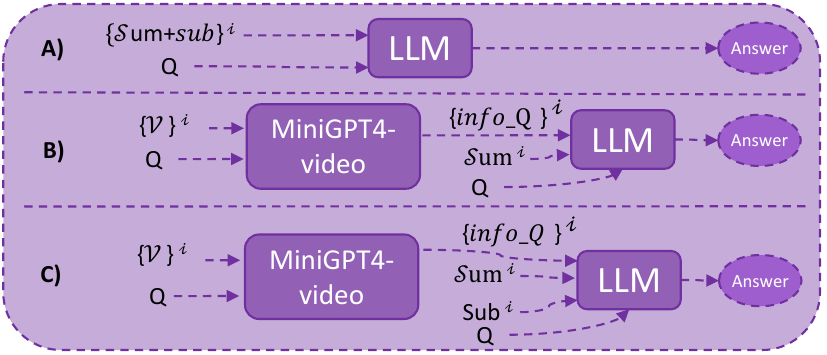}
    \end{center}
\end{minipage}
\begin{minipage}[c]{.28\linewidth}
    \begin{center}
    \scalebox{0.8}{
    \begin{tabular}{cc}
        \toprule
        \multirow{2}{*}{\textbf{{\makecell{Answer \\module}}}} & \textbf{Acc.} \\
        \- & \- \\
        \hline
        A) & 41.78\\
        B) & 27.6\\
        C) & 27.72\\
      \bottomrule
    \end{tabular}
    }
    \end{center}
\end{minipage}
\caption{Answer module ablation study on TVQA dataset, validation set.}
\label{tab_longvid_decoder_ablation}
\end{figure}

After getting the Top-K retrieved clips, the answer module is responsible to fuse the retrieved clips grounded by the question to produce the final answer.
To this end, several ways are studied, as shown in Figure \ref{tab_longvid_decoder_ablation}:
A) Feed directly the retrieved summaries $Sum$ and subtitles $Sub$ with the question to the LLM model to directly answer the question or say I don't know If the provided information not enough.
B) Feed the selected video clips $V$  and the question $Q$ to \shortModel~to generate a new information $info_Q$, which is grounded to the question. Then, feed the new information $info_Q$ with the general input summary $Sum$ and the question $Q$ to the LLM to produce the final answer.
C) Following the previous option,with also adding the original subtitles to the context.
The table in Figure \ref{tab_longvid_decoder_ablation} demonstrates that, option A is the best approach; feed the summaries and the subtitles directly to the LLM.
In contrast, when we feed the video clips $V$, the accuracy drops significantly, options B and C.
The reason behind this drop is the model hallucination, especially when the question is not related to the question, which leads to generate confusing information to the context $info_Q$.
Please refer to the supplementary materials for detailed examples of the model hallucination in options B and C.

\begin{figure}
    \centering
    \scalebox{0.6}{
    \includegraphics[width=1\linewidth]{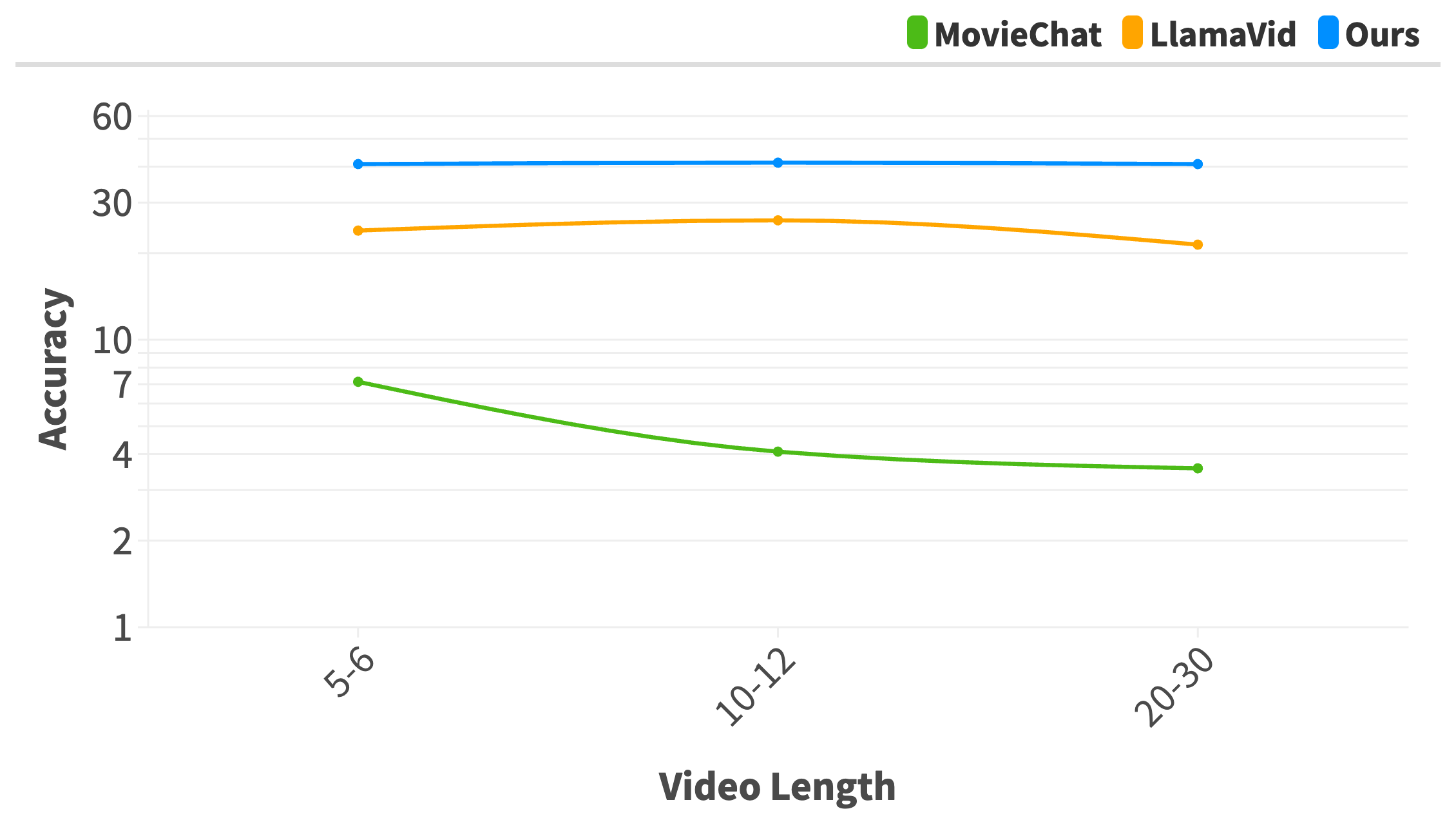}
    }
    \caption{Ablation study about the video length impact on 5\% of TVQA validation set. , video length in minutes}

    \label{fig_video_length_impact}
\end{figure}
To evaluate our framework's robustness with extended video lengths, we created three versions of the TVQA dataset by altering the aggregation window. This window compiles long videos from ground-truth short clips that include the answer to a question. Specifically, we combined 5, 10, and 20 clips to produce videos averaging 6, 12, and 24 minutes, respectively.
Figure \ref{fig_video_length_impact} illustrates that our framework maintains its robustness regardless of video length, with both retrieval performance and overall accuracy remaining consistent even as video duration increases.
These outcomes, detailed in Figure \ref{fig_video_length_impact}, are based on an analysis of 5\% of the TVQA validation set.

 

\begin{figure}[ht!]
    \captionsetup{font=scriptsize}
    \centering
    \scalebox{0.8}{
    \includegraphics[width=0.7\linewidth]{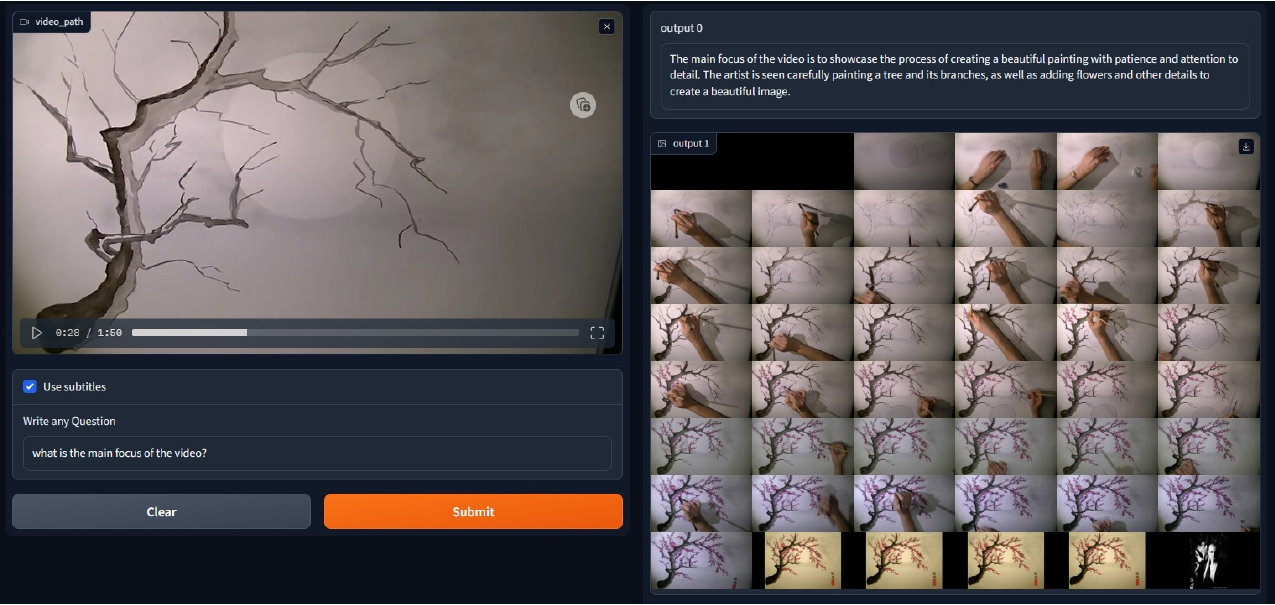}
    }
    \scalebox{0.8}{
    \includegraphics[width=0.7\linewidth]{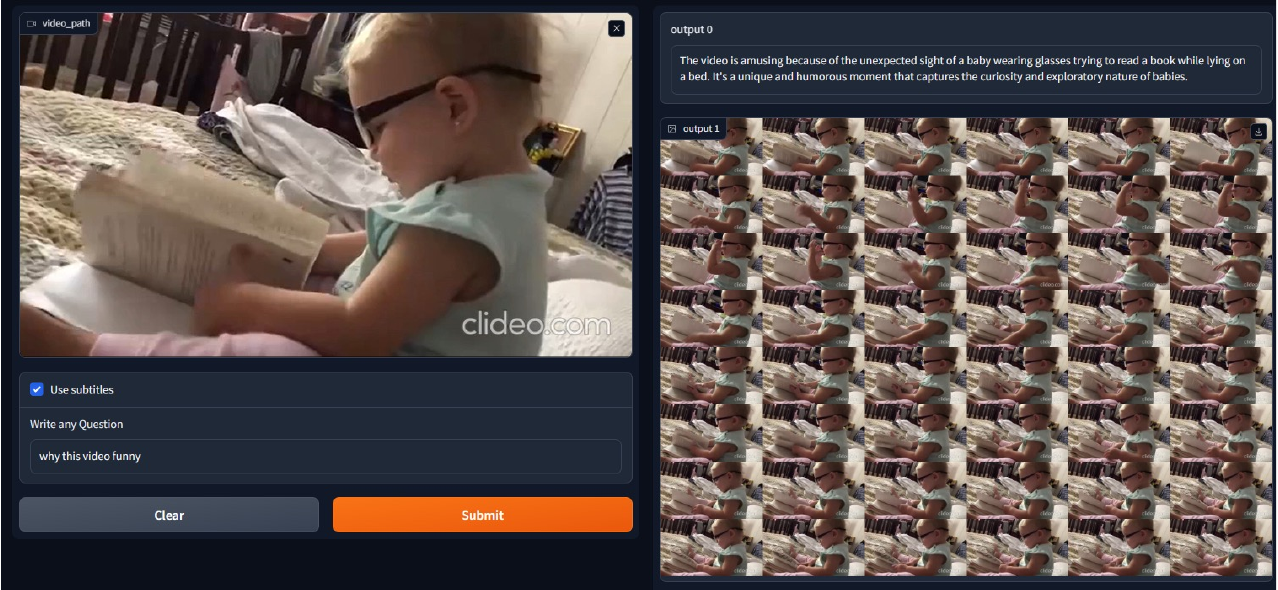}
    }
    \caption{MiniGPT4-video Qualitative results, demonstrating video understanding abilities; more qualitative results are provided in the supplementary.}
    \label{fig:qual1}
\end{figure}


\subsection{Comparison to State-Of-The-Art}

\subsubsection{Long Video Benchmarking}
We evaluate the efficacy of our proposed framework, {\papernameAbbrev}, across several well-established benchmarks, specifically the LLama-Vid~\cite{llamavid}, MovieChat\cite{moviechat}, Movie QA~\cite{movieqa}, and TVQA-Long~\cite{lei2019tvqa} datasets. To thoroughly examine our framework's capabilities, we analyze input modalities in two configurations: vision-only (V) and vision combined with input subtitles (V+T).

Our findings, detailed in Table \ref{tab:benchmark_longvideo}, indicate that our framework surpasses all existing long video baselines in the vision modality.We establish state-of-the-art (SOTA) performance on these challenging benchmarks. This achievement holds true even under an unfair comparison against LLama-Vid~\cite{llamavid}, which benefits from using the MovieNet dataset while training and these movies are in both LLama-vid \cite{llamavid} benchmark and Movie QA~\cite{movieqa}. Despite this advantage, our results significantly outperform the competition.

Incorporating both video frames and aligned subtitles into our model leads to an average performance boost of 8\% across the benchmarks. As highlighted in Table \ref{tab:benchmark_longvideo}, this enhanced approach enables us to outperform LLama-Vid~\cite{llamavid} on the TVQA benchmark, providing a fair comparison since LLama-Vid~\cite{llamavid} utilizes the other benchmarks during its training phase.

\subsubsection{Short Video Benchmarking}
On short-video understanding, we continue to secure state-of-the-art (SOTA) results, outperforming contemporaneous works, including LLama-Vid~\cite{llamavid}. To validate our framework's proficiency in short-video analysis, we conducted evaluations against current SOTA methodologies across an extensive suite of five benchmarks: Video ChatGPT, MSVD, MSRVTT, TGIF, and TVQA. These benchmarks collectively offer a comprehensive platform for assessing short-video comprehension capabilities, with five focusing on open-ended questions and TVQA featuring multiple-choice questions.

Our results, presented in Tables \ref{tab:benchmark_videochatgpt} and \ref{tab:benchmark_shortvideo}, demonstrate our framework's superiority over competing methods by a significant margin, affirming our considerable advancements across a varied and demanding collection of benchmarks. To thoroughly evaluate our approach, we devised two variations of our framework: one analyzing purely visual elements and another incorporating subtitles. The performance enhancements achieved with these models are noteworthy,  registering gains of 3.23\%, 2.03\%, 16.5\% and 23.59\% on the MSVD, MSRVTT, TGIF, and TVQA benchmarks respectively. This underscores our framework's ability to achieve SOTA results across the board, markedly elevating performance in the domain of short-video understanding. The visualization results of our method are shown in Fig.~\ref{fig:qual1}. We will show more visualization results in the appendix.

\tabcolsep=0.1cm
\begin{table}[t!]
\captionsetup{font=scriptsize}
\centering
\caption{Long video benchmarking results on four benchmarks: LLama-Vid, MovieChat, Movie QA, and our proposed TVQA-Long. The "V" modality indicates the use of video frames only, while "V+T" indicates the use of both video frames and subtitles. The dagger (†) symbol denotes methods that used the benchmark during training, implying an unfair comparison.}

\scalebox{0.8}{
\begin{tabular}{lcccccccccccc}
    \toprule
    \multicolumn{1}{c}{\multirow{3}{*}{Method}} & \- & \multicolumn{1}{c}{\multirow{3}{*}{Modalities}} & \- &  \multicolumn{4}{c}{Open Ednded Questions} & \- &  \multicolumn{4}{c}{MCQ} \\
    \cmidrule(r){5-9}
    \cmidrule(r){10-13}
    \- & \- & \- & \- & \multicolumn{2}{c}{LLama-Vid}\cite{llamavid} & \multicolumn{2}{c}{MovieChat}\cite{moviechat} & \- & \multicolumn{2}{c}{Movie QA} \cite{movieqa}& \multicolumn{2}{c}{TVQA-Long} \\
    \- & \- & \- & \- & Acc.↑ & Score↑ & Acc.↑ & Score↑ & \- & Acc.↑ & Score↑ & Acc.↑ & Score↑ \\
    \cmidrule(r){1-13}
    LLaMA-VID~\cite{llamavid}     & \- & V & \- &20.68 & \textbf{2.41}& 53.2& 3.81& \- & 24.42&2.19 & 24.63&2.16 \\
    MovieChat~\cite{moviechat}     & \- & V & \- &11.71 &1.45 & NA& NA& \- &16.18 &1.68 & 5.0& 0.86\\
    \rowcolor{blue!15} \textbf{Ours}   &\- & V & \- & \textbf{23.09}&2.19 & \textbf{67.6}&\textbf{ 4.23}& \- & \textbf{28.49}& \textbf{2.8}& \textbf{28.61}& \textbf{2.78}\\
    \hline
    LLaMA-VID ~\cite{llamavid} & \- & V+T & \- &41.4$^\dagger$  & 3.07$^\dagger$&NA & NA& \- &37.65$^\dagger$ &3.03$^\dagger$ &26.81 &2.21 \\
    \rowcolor{blue!15} \textbf{Ours}   &\- & V+T & \- &31.49 &2.43 & NA&NA & \- &35.24 & \textbf{3.1}& \textbf{41.78}&\textbf{3.21} \\
  \bottomrule
\end{tabular}
}
\label{tab:benchmark_longvideo}
\end{table}

\tabcolsep=0.1cm
\begin{table}[t!]
\captionsetup{font=scriptsize}
\centering
\caption{Qualitative results on Video-ChatGPT benchmark.}
\scalebox{0.8}{
\begin{tabular}{lcccccccc}
    \toprule
    \multicolumn{1}{c}{\multirow{3}{*}{\textbf{Method}}} & \- & \multicolumn{1}{c}{\multirow{3}{*}{\makecell{\textbf{Using} \\ \textbf{Subtitles}}}} & \- &  \multicolumn{5}{c}{\textbf{Video ChatGPT}}\\
    \cmidrule(r){5-9}
    \- & \- & \- & \- & \multirow{2}{*}{\makecell{Information\\Correctness}} & \multirow{2}{*}{\makecell{Detailed\\Orientation}}  &  \multirow{2}{*}{\makecell{Contextual\\Understanding}}  &  \multirow{2}{*}{\makecell{Temporal\\Understanding}} &  \multirow{2}{*}{Consistency}\\
    \- & \- & \- & \- & \-  & \-  &  \-  &   \-  &  \- \\

    \cmidrule(r){1-9}
    LLaMA Adapter~\cite{llama-adapter}    &\- & \xmark & \- & 2.03  & 2.32 & 2.30 & 1.98  &  2.15\\
    Video LLaMA~\cite{video-llama}  & \- & \xmark & \- &  1.96 & 2.18 & 2.16 & 1.82  & 1.79 \\
    Video-ChatGPT~\cite{videochatgpt}  &\- & \xmark & \- &  2.40 & 2.52 & 2.62 &  1.98 & 2.37 \\
    BT-Adapter-7B ~\cite{liu2023all}   & \- & \xmark & \- &  2.68 & 2.69 & 3.27 & 2.34  & 2.46 \\ 
    LLaMA-VID-7B~\cite{llamavid}     & \- & \xmark & \- &  2.96 & 3.00 & 3.53 &  2.46 & 2.51 \\
    \rowcolor{blue!15} \textbf{Ours-7B}   &\- & \xmark & \- &  2.93 & 2.97 & 3.45 & \textbf{2.47}  & \textbf{2.60} \\
    \hline
    Video Chat ~\cite{li2024videochat}  &\- & \cmark & \- & 2.23  & 2.50 & 2.53 & 1.94  & 2.24 \\
    \rowcolor{blue!15} \textbf{Ours-7B}   &\- & \cmark & \- &  \textbf{3.08} & \textbf{3.02} &\textbf{ 3.57} &  \textbf{2.65} & \textbf{2.67} \\
  \bottomrule
\end{tabular}
}
\label{tab:benchmark_videochatgpt}
\end{table}

\tabcolsep=0.1cm
\begin{table}[t!]
\captionsetup{font=small}
\centering
\caption{Short video benchmarking results on MSVD, MSRVTT, TGIF, ActivityNet and TVQA.}
\scalebox{0.8}{
\begin{tabular}{lccccccccccccccccc}
    \toprule
    \multicolumn{1}{c}{\multirow{3}{*}{\textbf{Method}}} & \- & \multicolumn{1}{c}{\multirow{3}{*}{\makecell{\textbf{Using} \\ \textbf{Subtitles}}}} & \- &  \multicolumn{8}{c}{\textbf{Open Ended Questions}} & \- &\multicolumn{1}{c}{\textbf{MCQ}}\\
    \cmidrule(r){5-12}
    \cmidrule(r){14-14}
    \- & \- & \- & \- & \multicolumn{2}{c}{MSVD} &\multicolumn{2}{c}{MSRVTT} &\multicolumn{2}{c}{TGIF} &\multicolumn{2}{c}{ActivityNet}  & \- &\multicolumn{1}{c}{TVQA}\\
    \cmidrule(r){5-6}
    \cmidrule(r){7-8}
    \cmidrule(r){9-10}
    \cmidrule(r){11-12}
    \cmidrule(r){14-14}
    \- & \- & \- & \- & \- Acc.↑ & Score↑ & Acc.↑ & Score↑ &  Acc.↑ & Score↑ &  Acc.↑ & Score↑ & \- &  Acc.↑ \\
    \cmidrule(r){1-4}
    \cmidrule(r){5-6}
    \cmidrule(r){7-8}
    \cmidrule(r){9-10}
    \cmidrule(r){11-12}
    \cmidrule(r){14-14}
    FrozenBiLM \cite{yang2022zeroshot}       &\- & \xmark & \- & 32.2  & --  & 16.8& -- & 41  & -- & 24.7  & -- & \- & 29.7 \\
    LLaMA Adapter~\cite{llama-adapter}    &\- & \xmark & \- & 54.9  & 3.1  & 43.8&2.7  & --  & -- & 34.2  & 2.7 &  \- &  --\\
    Video LLaMA~\cite{video-llama}  & \- & \xmark & \- & 51.6  & 2.5  & 29& 1.8 & --  &  --& 12.4  & 1.1 &  \- & -- \\
    Video Chat ~\cite{li2024videochat}  &\- & \xmark & \- & 56.3  &  2.8 & 45& 2.5 & 34.4  & 2.3 & 26.5  & 2.2 &  \- & -- \\
    
    Video-ChatGPT~\cite{videochatgpt}  &\- & \xmark & \- &  64.9 & 3.3  & 49.3& 2.8 & 51.4  & 3.0 & 35.2  & 2.7 & \- & 23.35 \\
    BT-Adapter-7B ~\cite{liu2023all}   & \- & \xmark & \- & 67.7  & 3.7  & 57& 3.2 & --  & -- & 45.7  & 3.2 & \- &  --\\ 
    LLaMA-VID-7B~\cite{llamavid}     & \- & \xmark & \- &  69.7 & 3.7  & 57.7& 3.2 & --  & -- & \textbf{ 47.4} & 3.3 &  \- & --  \\
    \rowcolor{blue!15} \textbf{Ours-7B}   &\- & \xmark & \- &  72.93 & 3.84  & \textbf{58.83}& \textbf{3.29} &  67.9 &3.71 & 45.6  & 3.2 &  \- &  36.45\\
    \hline
    \rowcolor{blue!15} \textbf{Ours-7B}   &\- & \cmark & \- &  N/A & N/A  & \textbf{59.73}& \textbf{3.3} & N/A & N/A  &  46.3 & \textbf{3.4} &  \- &  \textbf{46.94}\\
  \bottomrule
\end{tabular}
}
\label{tab:benchmark_shortvideo}
\end{table}

\section{Conclusion}

In this paper, we identified the main challenges of the current video-centric LLMs to process long videos. Based on the analyses, we introduced the \longModel~method, which eases the \textit{noise and redundancy} challenge and \textit{computational and memory} challenge. \longModel~introduces a retrieval approach that focuses on top-k relevant clips, allowing efficient processing of videos of any length. In contrast, most of the previous models can only process minutes-long videos. We developed \shortModel, which enhances video content interpretation from single to multiple frames, significantly improving video understanding. This model serves both as a part of \longModel~for long video summarization and as a standalone model for short video tasks. 
Our \longModel~achieves state-of-the-art results in long video understanding across four benchmarks with only the vision content and achieved SOTA in with vision and subtitles in zeroshot evaluation on TVQA. Notably, in the proposed TVQA-long benchmark, we outperformed the previous method by 14.94\%. Our \shortModel~also exceeds performance standards in short video benchmarks. We hope our proposed \longModel method and the TVQA-long benchmark can benefit future research in the long video understanding.

%
%
\clearpage
\bibliographystyle{splncs04}
\bibliography{main}
\clearpage
\renewcommand*{\thesection}{\Alph{section}}

\noindent{The supplementary material provides:}
\begin{itemize}
  \item Section~\ref{section:topk}: Ablation on different values of $k$ in Top-$k$ retrieval.
  \item Section~\ref{section:hallucination}: Model Hallucination problem.
  \item Section~\ref{sec_Minigpt4_video_in_text_tasks}: Video LLM in text tasks.
  \item Section~\ref{section:length}: Video length robustness.
  \item Section~\ref{section:prompt}: Prompt details.
  \item Section~\ref{section:detail}: Implementation details.
  \item Section~\ref{section:qual}: Qualitative results.
\end{itemize}

\section{Top K Effect}
\label{section:topk}
In this section, we explore how performance through accuracy is affected by the value of k for top k neighbors for the retrieval design in Section 3 of the main paper. From Table~\ref{number_of_neighbours}, we can see that the Top 3 achieved the best results for the ``Vision + subtitles'' experiments. By employing the general model summary, we observed that the accuracy improved when incorporating information from various neighbors. However, when this information was excessively increased, such as including data from five neighbors, the accuracy declined due to the introduction of noise from numerous incorrect details unrelated to the question. This phenomenon is evident in the first four rows.

From row 5 to 8 we can see that the accuracy decreased by increasing the number of neighbours because the related information from the wrong clips distract the model.
We observe the same behavior in the ``Vision Only'' and ``Subtitle Only'' experiments.

\begin{table}[h]
    \centering
    \caption{Effect of the number of neighbors on TVQA.Where model summary is the summary generated by the video descriptor and the Q\_related\_info is the new summary that is related to the question}
    \label{number_of_neighbours}
    \scalebox{0.8}{
    \begin{tabular}{lcc}
        \toprule
        \textbf{Model Variations} & \textbf{GPT-4 Accuracy (\%)} & \textbf{GPT-4 Score}\\
        \midrule
        \multicolumn{3}{l}{\textbf{Vision + Subtitles}} \\
        Top 1 (Model Summary + Subtitles) & 40.66 & 3.17 \\
        Top 2 (Model Summary + Subtitles) & 40.89 & 3.20 \\
        Top 3 (Model Summary + Subtitles) & 41.78 & 3.21 \\
        Top 5 (Model Summary + Subtitles) & 40.12 & 3.01 \\
        Top 1 (Model Summary + Subtitles + Q\_related\_info) &29.00  & 2.75 \\
        Top 2 (Model Summary + Subtitles + Q\_related\_info) & 28.12 & 2.71 \\
        Top 3 (Model Summary + Subtitles + Q\_related\_info) & 27.72 &2.69  \\
        \midrule
        \multicolumn{3}{l}{\textbf{Vision Only}} \\
        Top 1 (Model Summary) & 26.97 & 2.77 \\ 
        Top 2 (Model Summary)  & 27.72 & 2.77 \\ 
        Top 3 (Model Summary) & 28.61 & 2.78 \\
        Top 5 (Model Summary) & 27.63 & 2.67 \\
        Top 1 (Model Summary + Q\_related\_Info) & 27.83 &2.62  \\
        Top 2 (Model Summary + Q\_related\_Info) & 26.45 & 2.63 \\
        Top 3 (Model Summary + Q\_related\_Info) & 26.59 & 2.61 \\
        \midrule
        \multicolumn{3}{l}{\textbf{Subtitles Only}} \\
        Top 1 (Subtitles) & 40.23 & 3.15 \\
        Top 2 (Subtitles) & 41.61 & 3.20 \\
        Top 3 (Subtitles) & 41.80 & 3.22 \\
        Top 5 (Subtitles) & 39.83 & 3.02 \\
        \bottomrule
    \end{tabular}
    }
\end{table}

\section{Model Hallucinations}
\label{section:hallucination}
The model hallucinates in our case when the VideoLLM is asked questions unrelated to the video, so the videoLLM generates incorrect information which misguides the answer module to answer the right answer.

After retrieving the Top-$k$ clips, our goal is to filter these clips to the single correct one. Theoretically, we could prompt each retrieved clip with the query and filter for which clip produces an answer. A common problem in generative models, we find that the model hallucinates and outputs an answer instead of stating it doesn't have the required information to answer. This issue particularly arises when the clips originate from the same episode. However, we do see that the videoLLM responding with its lack of information to answer the question if the clip is entirely unrelated to the question. 

For instance, in the multi-choice questions in TVQA, if the top three retrieved clips are guaranteed that one of them is the correct clip and the other two clips are incorrect, when using the VideoLLM with the wrong clip it will choose a wrong choice and when feeding the other wrong one, it will choose another wrong choice, and when using the correct clip, it may choose the correct choice based on the correct video content or may choose the wrong choice. In both cases the answer module will see the context information has three choices and this distracts it from answering correctly even if one of them is the correct answer as evidenced by the table 
\ref{model_hallucination}. the accuracy dropped by around 14 \% in the vision and subtitles and dropped by 2 \% in the vision only.





\begin{table}[h]
    \centering
    \caption{Effect of model hallucination.Where the model summary is the summary generated by the video descriptor and the Q\_related\_info is the new summary that is related to the question.
    }
    \label{model_hallucination}
    \scalebox{0.9}{
    \begin{tabular}{lcc}
        \toprule
        \textbf{Model Variations} & \textbf{GPT-4 Accuracy (\%)} & \textbf{GPT-4 Score} \\
        \midrule
        \multicolumn{3}{l}{\textbf{Vision + Subtitles}} \\
        Top 3 (Model Summary + Subtitles) & 41.78 & 3.21 \\
        Top 3 (Model Summary + Subtitles + Q\_related\_info) & 27.72 &2.69  \\
        \midrule
        \multicolumn{3}{l}{\textbf{Vision Only}} \\
        Top 3 (Model Summary) & 28.61 & 2.78 \\ 
        Top 3 (Model Summary + Q\_related\_Info) & 26.59 & 2.61 \\
        \bottomrule
    \end{tabular}
    }
\end{table}

\section{\shortModel~in Text Tasks}
\label{sec_Minigpt4_video_in_text_tasks}
Here, we will see how the fine-tuned version of Llama 2 (our \shortModel) performs compared to the original Llama2 in the text tasks.
We used \shortModel~as an answer module in the Goldfish system. We can tell from the table \ref{tab_Minigpt4_video_in_text_tasks} that \shortModel~has lost some text skills during vision tasks fine-tuning, so we decided to use the original Llama to get the best performance. 
\begin{table}[h]
    \centering
    \caption{Ablation about answer module LLM}
    \label{tab_Minigpt4_video_in_text_tasks}
    \scalebox{0.9}{
    \begin{tabular}{lcc}
        \toprule
        \textbf{Top 3 (Model Summary + Subtitles)} & \textbf{GPT-4 Accuracy (\%)} & \textbf{GPT-4 Score} \\
        \midrule
        Goldfish with \shortModel~as answer module &35.07 &2.93 \\
        \midrule
        Goldfish with original Llama2 as answer module & 41.78 & 3.21 \\
        \bottomrule
    \end{tabular}
    }
\end{table}

\section{Video Length Robustness.}
\label{section:length}

\begin{wraptable}{r}{0.3\textwidth}
\captionsetup{font=scriptsize}
\caption{Ablation study about the video length impact on 5\% of TVQA validation set.}
\label{tab_video_length_impact}
\centering
\scalebox{0.55}{
\begin{tabular}{ccc}
    \toprule
    Video Length & Retrieval Acc. & Overall Acc. \\
    \cmidrule(r){1-1}
    \cmidrule(r){2-3}
    5-6 Min  & 60.2 & 40.8 \\
    10-12 Min & 60.2 & 41.3 \\
    20-30 Min & 60.2 & 40.8 \\
\bottomrule
\end{tabular}
}
\end{wraptable}
To evaluate our framework's robustness with extended video lengths, we created three versions of the TVQA dataset by altering the aggregation window. This window compiles long videos from ground-truth short clips that include the answer to a question. Specifically, we combined 5, 10, and 20 clips to produce videos that average between 6, 12, and 24 minutes, respectively.
Table \ref{tab_video_length_impact} illustrates that our framework maintains its robustness regardless of video length, with both retrieval performance and overall accuracy remaining consistent even as video duration increases.
These results, detailed in Table \ref{tab_video_length_impact}, are based on an analysis of 5\% of the TVQA validation set.

\section{Prompts Details}
\label{section:prompt}

\subsubsection{Evaluation prompts.} We followed the same evaluation setting in videochatgpt\cite{videochatgpt}. The \{question\}, \{answer\}, and \{pred\} correspond to the question, the ground truth answer, and the model prediction, respectively, in the prompt. The \textbf{System prompt} is as follows:

{\color{darkgray}
You are an intelligent chatbot designed for evaluating the correctness of generative outputs for question-answer pairs.
Your task is to compare the predicted answer with the correct answer and determine if they match meaningfully. Here's how you can accomplish the task:\\
INSTRUCTIONS: 
\begin{itemize}
    \item Focus on the meaningful match between the predicted answer and the correct answer.
    \item Consider synonyms or paraphrases as valid matches.
    \item Evaluate the correctness of the prediction compared to the answer.
\end{itemize}
User prompt:\\
Please evaluate the following video-based question-answer pair:
\begin{flushleft}
    Question: \{question\} \\
    Correct Answer: \{answer\} \\
    Predicted Answer: \{pred\}
\end{flushleft}

Provide your evaluation only as a yes/no and score where the score is an integer value between 0 and 5, with 5 indicating the highest meaningful match. Please generate the response in the form of a Python dictionary string with keys `pred' and `score', where the value of `pred' is a string of `yes' or `no' and the value of `score' is an INTEGER, not STRING. DO NOT PROVIDE ANY OTHER OUTPUT TEXT OR EXPLANATION. Only provide the Python dictionary string. For example, your response should look like this: \{`pred': `yes', `score': 4.8\}. 
}

\subsubsection{Summary prompts.} 
Below is the summary prompt to obtain the vision summary of the clip:\\

{\color{darkgray}
Generate a description of this video. Pay close attention to the objects, actions, emotions portrayed in the video, providing a vivid description of key moments. Specify any visual cues or elements that stand out.
}

\subsubsection{Extract the related information prompt}:
In the multi-choice questions, we added the choice \textit{``I don't know''} as the fifth choice, and the \{question\} is a placeholder for the question itself in the prompt. The prompt is as follows:\\

{\color{darkgray}
From this video extract the related information to This multichioce question and provide an explaination for your answer and If you don't know the answer, say 'I DON'T KNOW' as option 5 because maybe the questoin is not related to the video content. the question is: \{question\} your answer:
}

\section{Implementation Details}
\label{section:detail}

Our models are trained with 4 A100 GPUs.
 The training process involved three distinct stages, with specific durations allocated to each. The initial stage focused on image-text training and spanned a period of two days. Subsequently, the second stage, dedicated to pre-training with video captions datasets, lasted one day, followed by the third stage, involving instruction tuning, which extended over three days. Throughout these stages, we maintained a batch size of 4 and utilized the AdamW optimizer in conjunction with a cosine learning rate scheduler, setting the learning rate to 1e-4.

 Our visual backbone consisted of the EVA-CLIP V1~\cite{sun2023eva} architecture, with the frozen weights. Notably, we trained the linear projection layer and performed efficient fine-tuning of the language model using LoRA~\cite{hu2021lora} (\textit{Low-Rank Adaptation}). Specifically, we fine-tuned the $W_q$ and $W_v$ components with a rank (\textit{r}) of 64 and a LoRA-alpha value equal 16. The entire model was trained with a consistent image resolution of $224\times224$ pixels, ensuring uniformity across all stages. 

\section{Qualitative Results}
\label{section:qual}
\subsection{Long Video}
\begin{figure}[t!]
    \centering
    \includegraphics[width=1.0\linewidth]{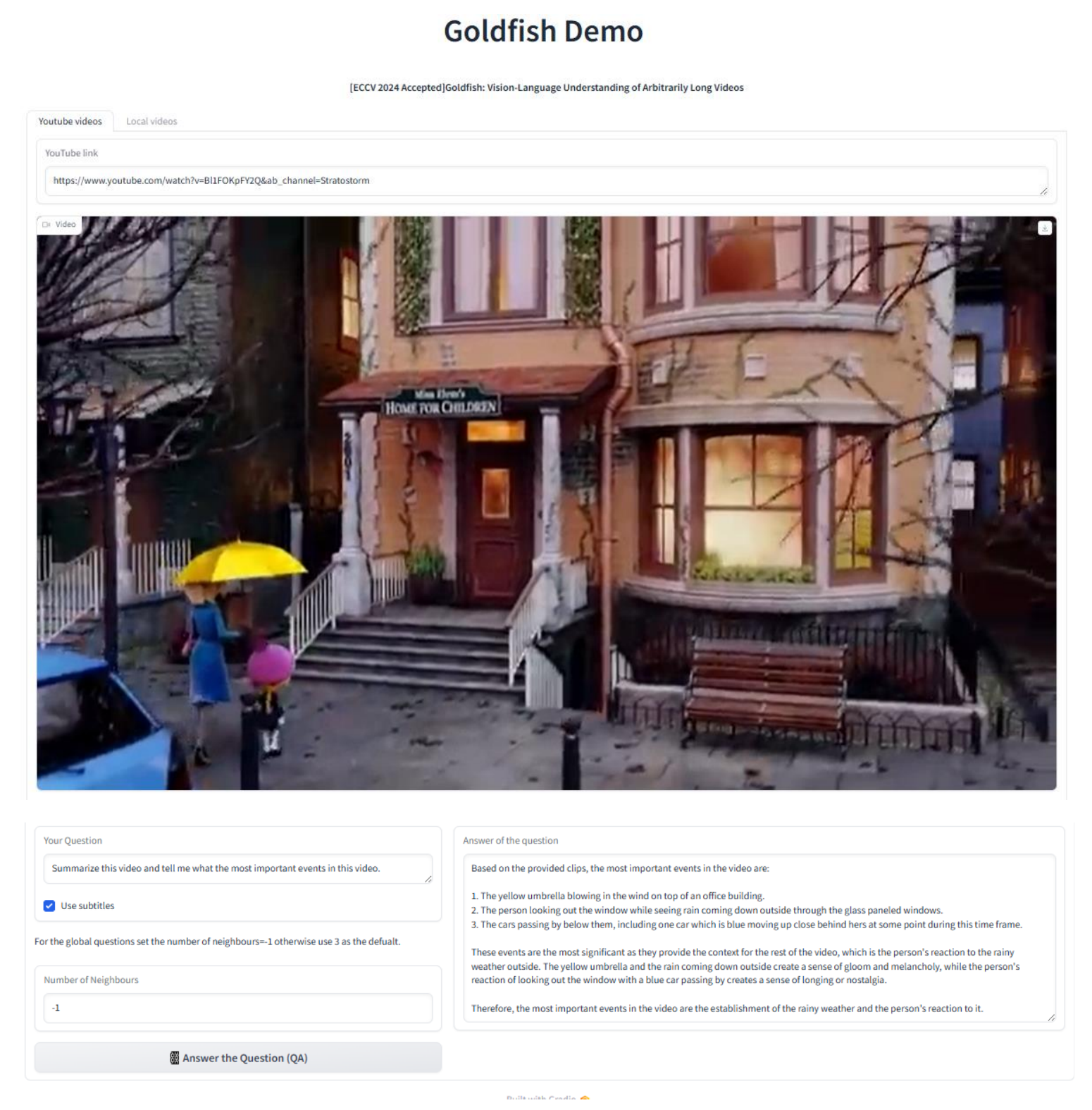}
    \captionsetup{font=scriptsize}
    \caption{Qualitative result of Goldfish model for long video understanding. \href{https://www.youtube.com/watch?v=Bl1FOKpFY2Q&ab_channel=Stratostorm}{[link]}.}
    \label{fig:long_1}
\end{figure}

\begin{figure}[t!]
    \centering
    \includegraphics[width=1.0\linewidth]{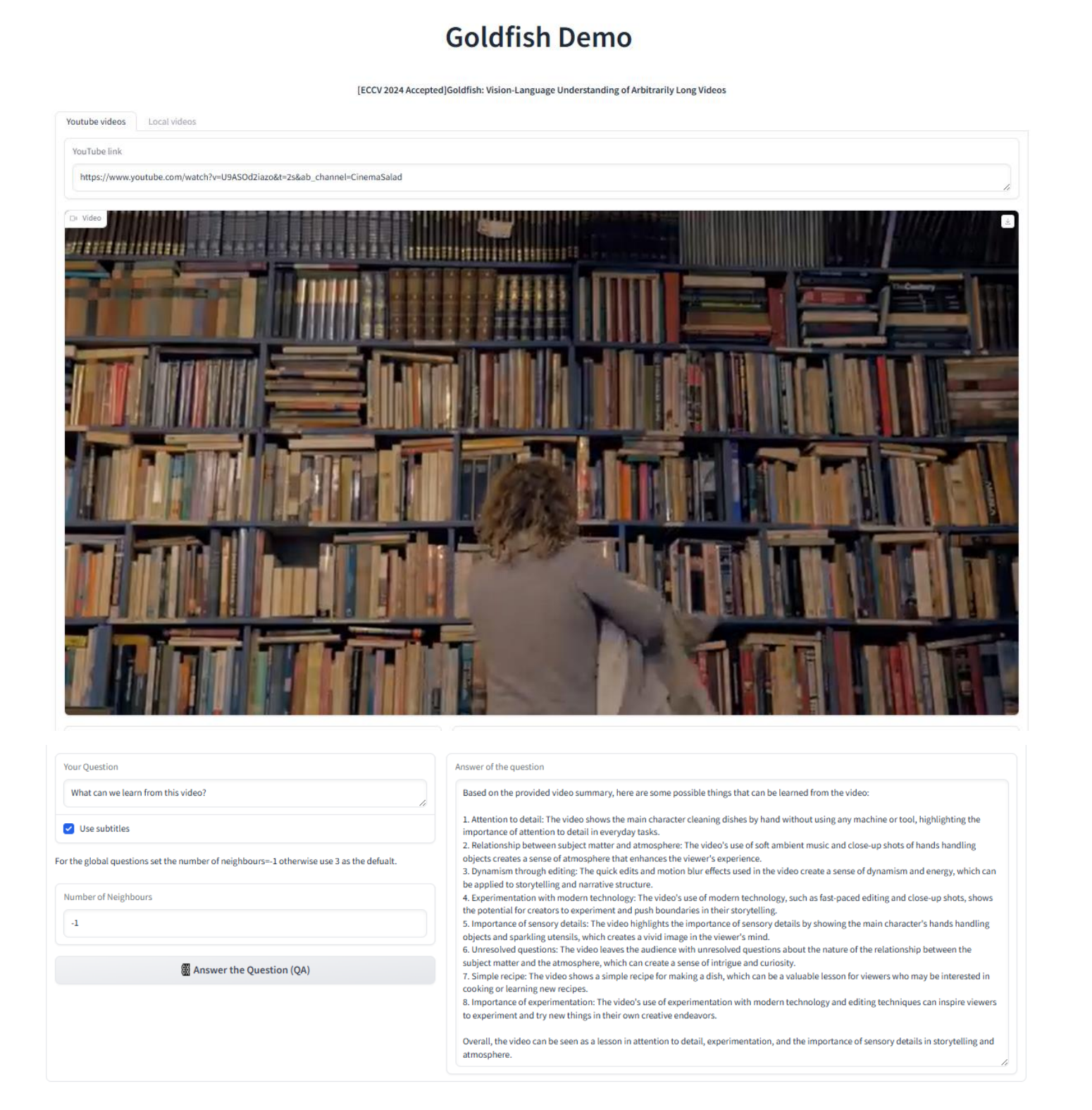}
    \captionsetup{font=scriptsize}
    \caption{Qualitative result of Goldfish model for long video understanding. \href{https://www.youtube.com/watch?v=U9ASOd2iazo&t=2s&ab_channel=CinemaSalad}{[link]}.}
    \label{fig:long_2}
\end{figure}

Fig \ref{fig:long_1} and Fig \ref{fig:long_2}\ shows one example of the goldfish demo.
Please refer to this \href{https://1drv.ms/u/s!ApW05sOkCBBda4QP8kNVwa9WbFE?e=XnOdJf}{link} for more qualitative video demos.

\subsection{Short Video}

\begin{figure}[t!]
    \centering
    \includegraphics[width=1.0\linewidth]{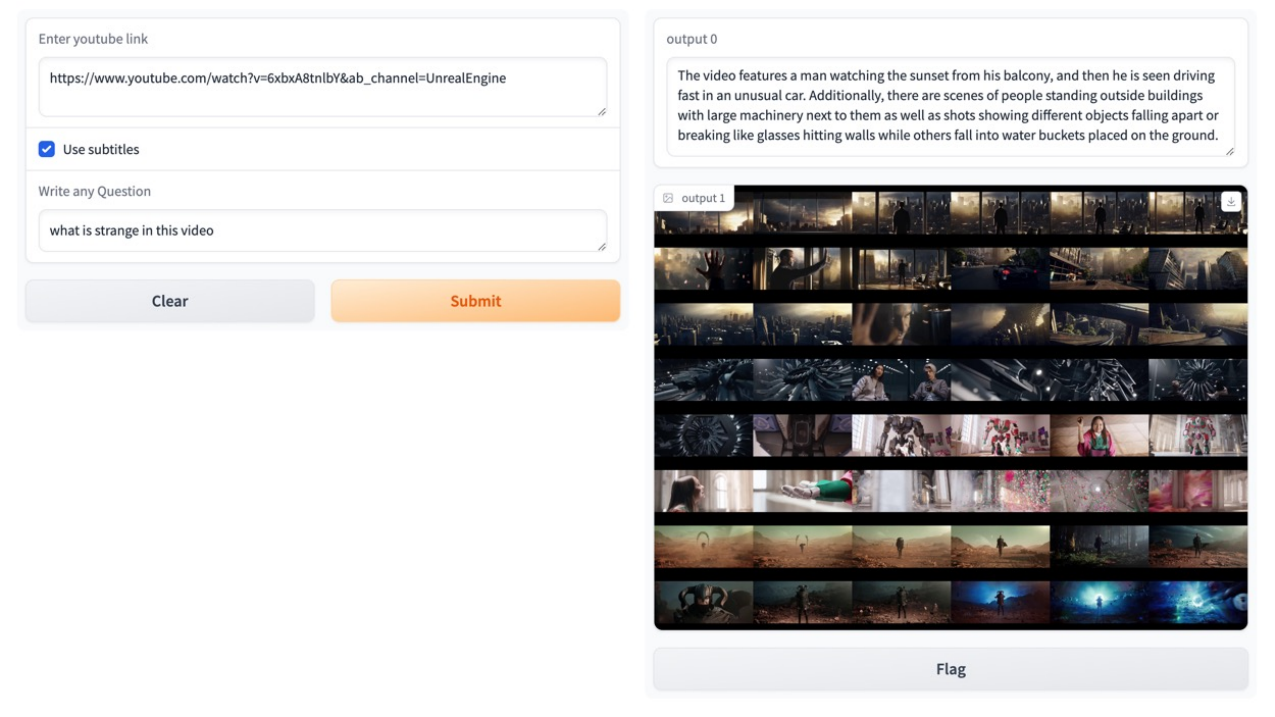}
    \captionsetup{font=scriptsize}
    \caption{Qualitative result of short video understanding with in-the-wild video \href{https://www.youtube.com/watch?v=6xbxA8tnlbY&ab_channel=UnrealEngine}{[link]}.}
    \label{fig:short1}
\end{figure}

\begin{figure}[t!]
    \centering
    \includegraphics[width=1.0\linewidth]{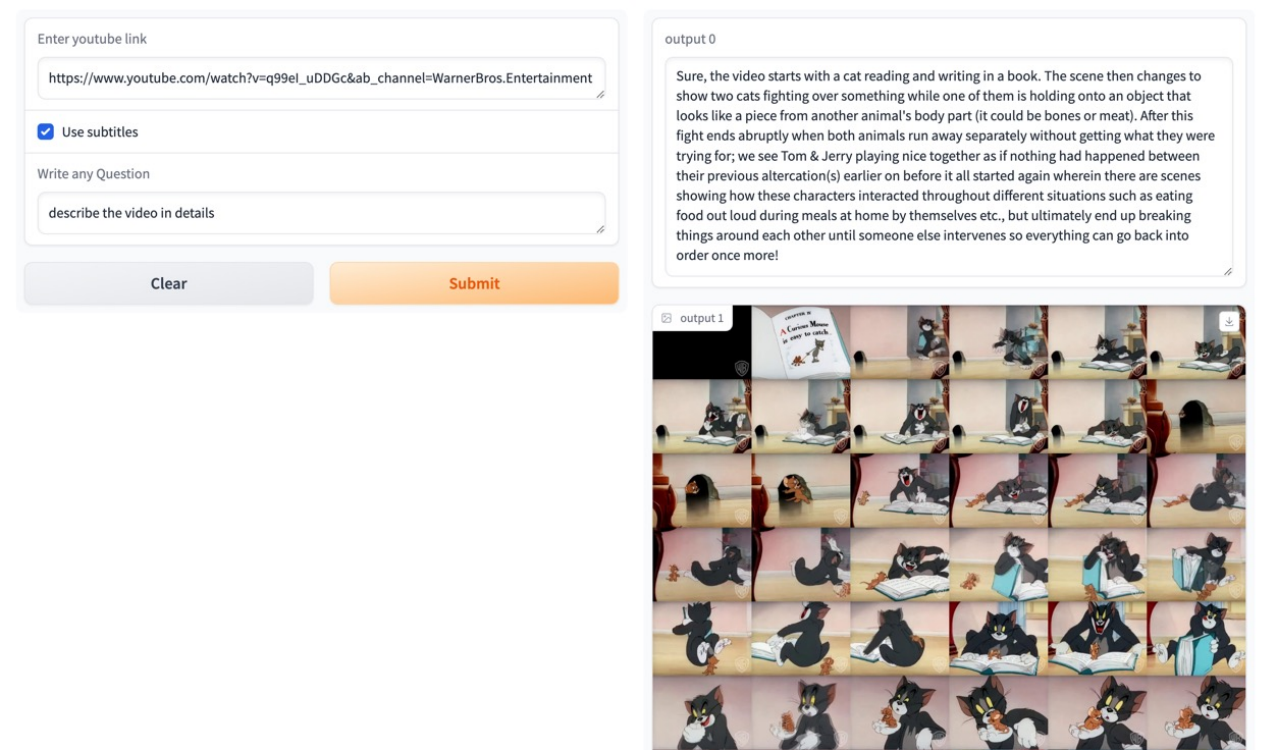}
    \captionsetup{font=scriptsize}
    \caption{Qualitative result of short video understanding with in-the-wild video \href{https://www.youtube.com/watch?v=q99eI\_uDDGc\&ab\_channel=WarnerBros.Entertainment}{[link]}.}
    \label{fig:short2}
\end{figure}

\cref{fig:short1,fig:short2} demonstrate qualitative results of our model MiniGPT4-video on in-the-wild online videos.


\begin{figure}[t!]
    \centering
    \includegraphics[width=1.0\linewidth]{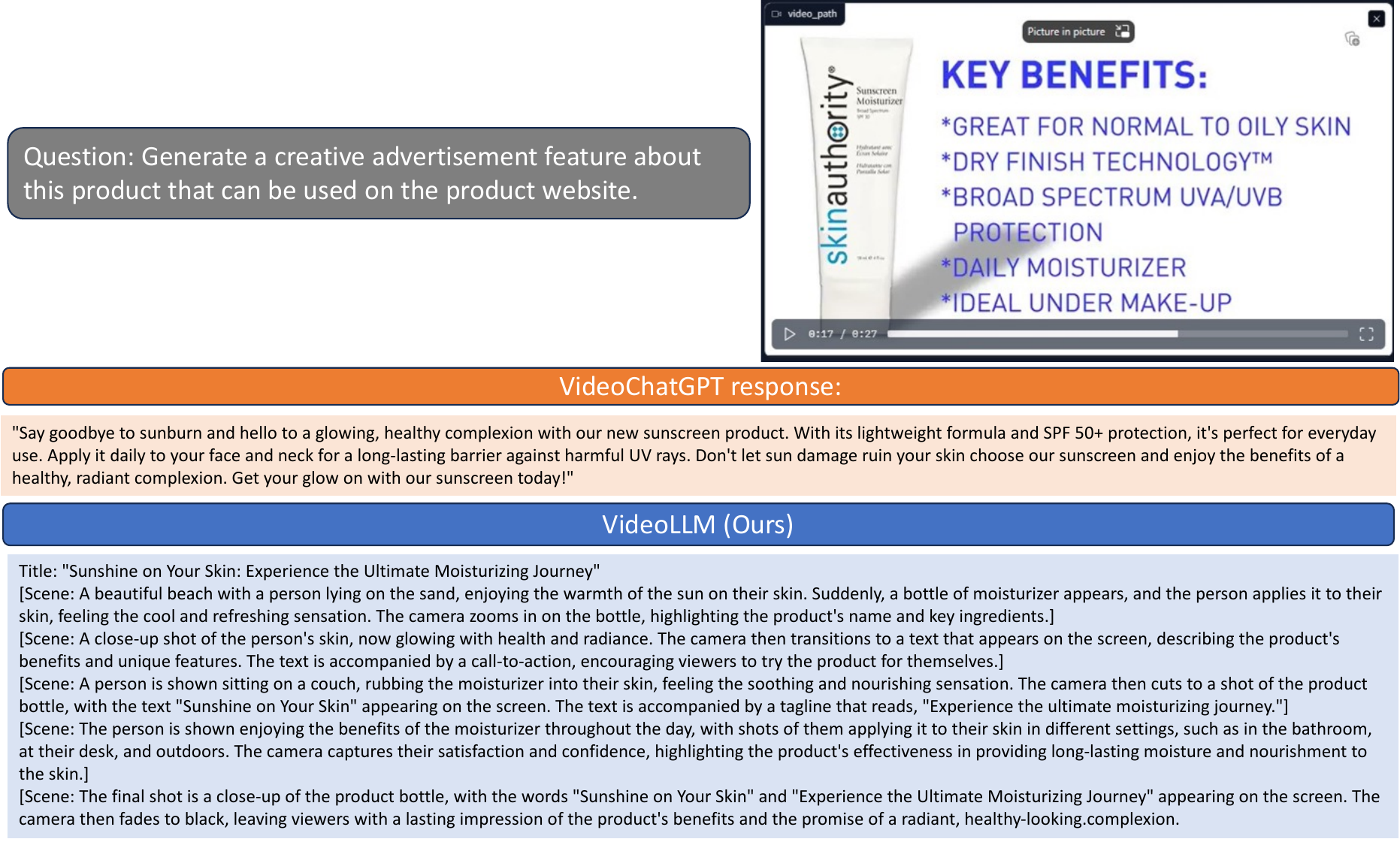}
    \captionsetup{font=scriptsize}
    \caption{\textbf{Short model qualitative results Ours vs VideoChatGPT~\cite{videochatgpt}.}}
    \label{fig:visionllm}
\end{figure}




\end{document}